\documentclass[letterpaper, 10 pt, conference]{ieeeconf}  % Comment this line out if you need a4paper

\IEEEoverridecommandlockouts                              % This command is only needed if 
\usepackage{graphics} % for pdf, bitmapped graphics files
\usepackage{epsfig} % for postscript graphics files
\usepackage{amsmath} % assumes amsmath package installed
\usepackage{amssymb}  % assumes amsmath package installed
\usepackage{hyperref}

\usepackage[isbn=false, backend=biber, style=ieee, bibstyle=ieee,
            sortcites=true, giveninits=true, backref=false]{biblatex}

\usepackage{booktabs}   
\usepackage{placeins}
\usepackage{pifont}
\newcommand{\yes}{\textcolor{ForestGreen}{\ding{51}}}
\newcommand{\no}{\textcolor{BrickRed}{\ding{55}}}

\usepackage[dvipsnames]{xcolor}

\title{\LARGE \bf
See to Reach, Feel to Grasp: Learning A Blind Grasp Reflex for Anthropomorphic Robotic Hands
}

\author{
Alexander Alexiev$^{1}$, Tzu-Yuan Lin$^{1}$, Sang Min Kim$^{1,2}$, Ho Jae Lee$^{1}$, Yonghyeon Lee$^{3,*}$ and Sangbae Kim$^{1,*}$% <-this % stops a space
\thanks{$^{1}$Department of Mechanical Engineering, Massachusetts Institute of Technology. $^{2}$Department of Electrical and Computer Engineering, Seoul National University. $^{3}$Department of Artificial Intelligence, Yonsei University. $^{*}$Yonghyeon Lee and Sangbae Kim are co-corresponding authors.
Emails: yonghyeon.lee@yonsei.ac.kr and sangbae@mit.edu.}
 }
\begin{document}

\maketitle
\thispagestyle{empty}
\pagestyle{empty}
\renewcommand{\figureautorefname}{Fig.}
\renewcommand{\equationautorefname}{Eq.}
\renewcommand{\sectionautorefname}{Section}
\renewcommand{\subsectionautorefname}{Section}

\newcommand{\comment}[1]{{\color{blue}Comment: #1}}
\newcommand{\todo}[1]{{\color{green}todo: #1}}
\newcommand{\hojae}[1]{{\color{cyan}Ho Jae: #1}}
\newcommand{\smkim}[1]{{\color{orange}Sang Min: #1}}
\newcommand{\ylee}[1]{{\color{red}Y. Lee: #1}}
\newcommand{\mitmanipulator}{a manipulator platform}
\newcommand{\graspscore}{\gamma}

%%%%%%%%%%%%%%%%%%%%%%%%%%%%%%%%%%%%%%%%%%%%%%%%%%%%%%%%%%%%%%%%%%%%%%%%%%%%%%%%
\begin{abstract}
% Seeing an object helps a robot reach it—but must the hand keep seeing it to grasp? We present a modular dexterous grasping architecture that separates global arm motion from local contact control. An independently controlled arm guides the hand toward the object, while a reinforcement learning policy grasps and stabilizes the object using only hand proprioception feedback, without images, object poses, or geometric observations. 

% In this work we study if a proprioceptive only hand policy is sufficient for grasping diverse objects 
% Seeing an object helps a robot reach it, but must the hand continue seeing it to grasp? 

In this work we study if a robotic hand using proprioception alone can grasp diverse objects with no visual observation. We present a modular dexterous grasping architecture that separates global arm motion from local contact control. An independently controlled arm guides the hand toward the object, while a reinforcement learning policy grasps and stabilizes it using only hand proprioceptive feedback. We call this \textit{a blind grasp reflex}: grasping without images, object poses, or geometric observations. A learned stable-grasp score determines when the object is securely held, allowing the arm to begin post-grasp manipulation. This separation makes grasping a reusable hand-level skill that can be combined with independently designed arm controllers for various manipulation tasks. Experiments in simulation and on hardware demonstrate robust blind grasping across diverse objects and seamless composition with a range of arm controllers. Moreover, despite never observing contact geometry, the learned grasp score closely aligns with an independent physics-based measure of grasp stability. The resulting approach follows a simple principle: see to reach, feel to grasp.
Project page: \href{https://blindgraspreflex.github.io}{blindgraspreflex.github.io}. 
\end{abstract}

\section{Introduction}

Dexterous grasping is fundamentally a contact-rich feedback problem. Parallel-jaw
grippers can often secure an object by squeezing it between two broad frictional
surfaces. Dexterous hands establish multiple localized contacts whose configuration and force distribution determine grasp quality ~\cite{ferrari1992planning, roa2015grasp, li2023frogger}. 
Explicitly estimating and modeling these interactions is difficult, particularly without dense
tactile sensing across the hand~\cite{shawcortez2019robust, calandra2018more,
pan2026beyondbinary}.

Many grasping systems simplify this problem by separating reaching from closing.
Given an object observation, they generate a target grasp configuration, plan a
collision-free approach to a pre-grasp pose, and close the fingers toward a
prescribed final configuration~\cite{li2023frogger, xu2023unidexgrasp}. This decomposition works well when perception and reaching are
accurate, but becomes brittle under uncertainty. Once contact occurs, closing toward a fixed configuration offers
limited flexibility to adjust contacts in response to residual reaching
errors, perception uncertainty, or
disturbances.

Reinforcement learning (RL) provides a way to learn such feedback behavior through interaction~\cite{qin2022dexpoint, wan2023unidexgrasp2,
zhang2025robustdexgrasp}. Many learned grasping policies, however, condition contact
control on global visual or geometric information, such as
images~\cite{singh2024dextrahrgb}, point clouds~\cite{zhang2025robustdexgrasp}, object
poses~\cite{lee2026reactive}, or shape features~\cite{xu2023unidexgrasp}. These
observations help locate the object and guide the approach. However, conditioning local
contact control on global geometry may introduce dependencies on object shape, pose,
and scene configuration that impede generalization. 

% In this work, we investigate an alternative approach, where the grasping acquisition and maintenance is formualted as a hand-local feedback skill, with an explicit interface for coordinating independently designed arm controllers.

\begin{figure}[!t]
    \centering
    \includegraphics[width=0.85\linewidth]{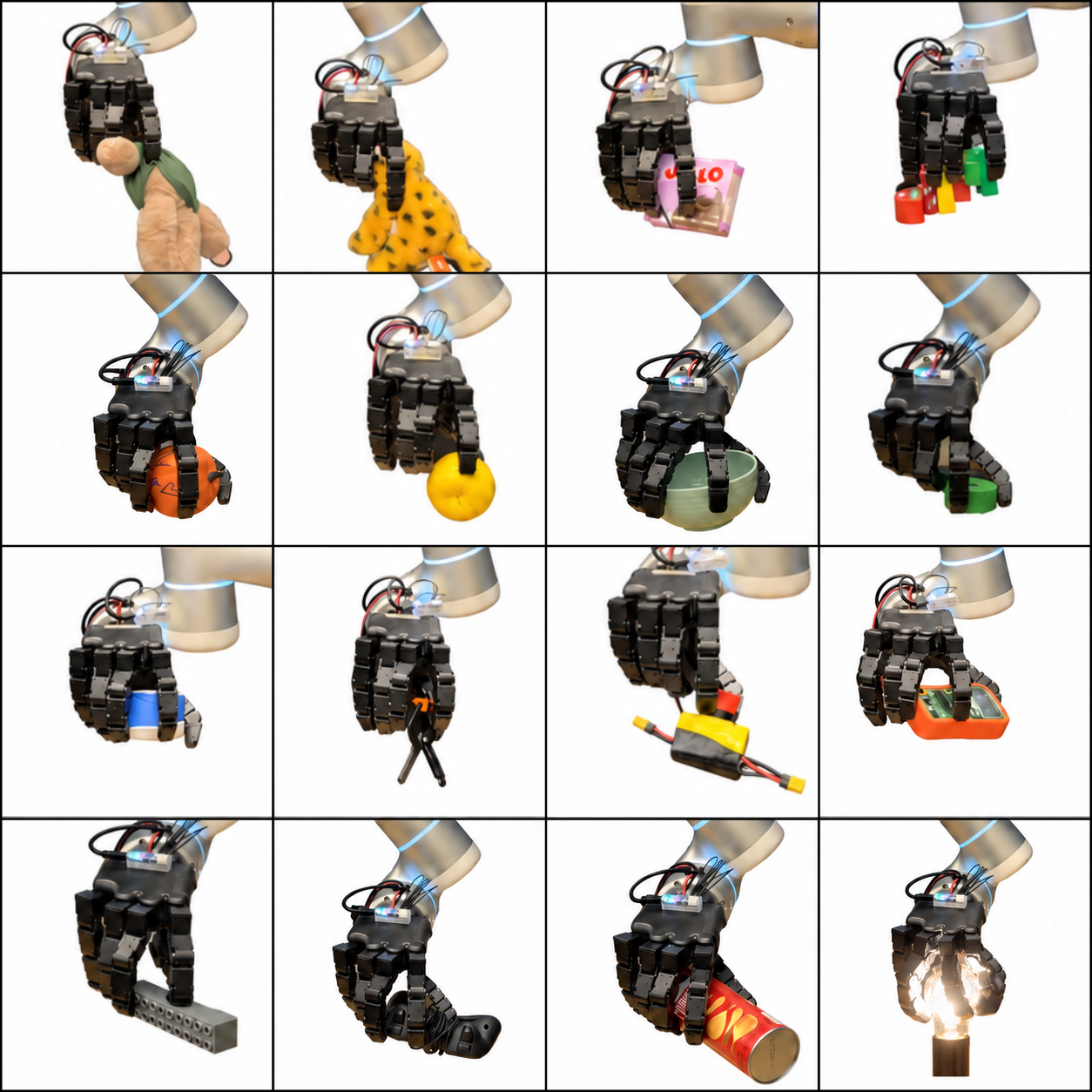}
    \caption{
    The proposed method can grasp a diverse set of real world objects blindly without visual observations and priors.
    }
    \label{fig:grid}
    \vspace{-6mm}
\end{figure}

Human grasping motivates a division of responsibility: ``see to
reach, feel to grasp.'' We can bring a hand toward an approximately
located object and complete the grasp without continuously
observing it~\cite{johansson2009coding, westling1984factors}.
We study this separation through \textit{blind grasping}, in which
grasp control operates without visual input.
Prior work has combined vision-guided pre-grasp positioning with
blind reactive hand control for functional
grasping~\cite{agarwal2023dexterous}. However, its
reported hand policy conditions on a target end-effector pose,
retaining an explicit input dependency on the arm's motion target.
In contrast, we investigate whether grasp acquisition and maintenance can
instead rely on hand-local observations, without arm-state or
end-effector-target inputs.

% Vision guides the approach, while proprioceptive feedback helps
% the fingers establish contact~\cite{lee2026whylook}, compensate for reaching
% errors~\cite{luo2026blind}, and maintain a stable grasp~\cite{shawcortez2019robust}.

To this end, we propose an architecture that assigns global geometric reasoning to the arm controller and local contact control to a learned hand policy. The arm brings the hand toward a feasible grasp, while the hand policy uses only local sensory feedback to grasp the object. The hand need not wait for the arm to reach a precise pre-grasp pose: it can respond as contacts emerge and accommodate residual reaching errors through feedback. Excluding global visual and geometric observations from the hand policy is intended to support reuse of the same stabilization behavior across objects, poses, and scene configurations.

Coordinating adaptive hand control with the arm also requires
deciding when to begin post-grasp manipulation. As contacts develop differently
across grasp attempts, a fixed transition time may initiate arm
motion before the grasp is ready or impose unnecessary waiting.
Prior functional grasping work~\cite{agarwal2023dexterous} learns
blind reactive finger control under a fixed training-time
arm-motion schedule, but does not describe a learned hand-to-arm
signal for coordinating the transition to post-grasp motion. We therefore learn a grasp score from interaction experience. Once the grasp score returns as stable, the arm proceeds with the desired manipulation motion while the hand policy remains active to maintain contact and stability.

% Continuous contact control also requires a criterion for grasp completion. Because the hand does not converge to a prescribed final configuration, reaching a fixed joint target cannot determine whether the object is securely held. We therefore learn a grasp readiness score from interaction experience. Once the grasp score returns as stable, the arm proceeds with the desired manipulation motion while the hand policy remains active to maintain contact and stability.

We evaluate our approach on a 20-DoF anthropomorphic hand
mounted on a 7-DoF torque-controlled arm. In simulation, the blind policy grasps 96\% of
YCB \cite{calli2015ycb} and 95\% of 3028 randomly sampled GraspXL \cite{zhang2024graspxl} objects in simulation. Although trained
without stability labels, the learned grasp score correlates with a force closure Ferrari--Canny
$\epsilon$ margin \cite{ferrari1992planning} computed from contacts the policy never observes
($\bar{\rho} = +0.77$). On hardware we grasp 29 different real-world objects and pair the frozen grasping policy with three arm controllers unseen during training to accomplish different manipulation tasks including VLM based general pick and place, screwing in a lightbulb, and dynamic grasping.

% The resulting architecture integrates reaching, contact acquisition, and stabilization: global geometry guides the approach, local feedback establishes and maintains the grasp, and a learned stability estimate determines when to begin manipulation.

The main contributions of this work include: 
\begin{itemize}
    \item We learn a hand-local policy that grasps and stabilize using joint encoders as its only sensory input. The
    policy can be paired with independently designed arm controllers.
    
    \item We introduce a bidirectional arm-hand interface. The arm activates grasping, while a learned grasp
    score signals when to begin post-grasp manipulation.

    \item We evaluate grasping robustness and present qualitative
    hardware demonstrations on diverse objects, without providing
    object geometry to the deployed hand policy.
\end{itemize}

\section{Related Works}
\label{sec:related}

\subsection{Vision-Based Dexterous Grasping}

Learning-based dexterous grasping methods commonly use visual or object-state observations to guide grasp acquisition. DexPoint~\cite{qin2022dexpoint} and the UniDexGrasp~\cite{xu2023unidexgrasp,wan2023unidexgrasp2} family condition policies on point clouds or object state to support grasping across diverse objects. DextrAH-G~\cite{lum2024dextrah} and DextrAH-RGB~\cite{singh2024dextrahrgb} distill privileged teachers into depth- or RGB-driven student policies that jointly control the arm and hand. These approaches integrate perception, reaching, and contact acquisition within a learned control framework.

Other work improves how geometry is represented for grasping. RobustDexGrasp~\cite{zhang2025robustdexgrasp} uses a hand-centric distance representation derived from single-view point clouds, while related approaches address grasping in cluttered scenes~\cite{wang2025clutterdexgrasp}. Expressing geometry relative to the hand provides a useful alternative to global object representations, but the policy still depends on externally perceived geometry. Likewise, joint arm--hand control can coordinate approach and grasp formation effectively, yet makes the hand behavior less straightforward to reuse with an independently designed reaching controller.

Our work differs in both observation design and control scope. The learned policy controls only the hand and grasps and stabilizes the object from local sensory history, without visual or object-geometric observations. Arm motion is handled independently, allowing the grasp policy to be composed with separate reaching and manipulation controllers. This separation complements vision-based methods: visual perception can still guide the arm, while the hand adapts to contact through local feedback.

\subsection{Blind and Proprioceptive Grasping}

Grasping without vision has largely relied on mechanical compliance or dedicated tactile sensing.
Compliant underactuated hands adapt to uncertainty in object shape and position through passive joint compliance and tendon differentials, achieving enveloping grasps with minimal sensing~\cite{odhner2014ihy,deimel2016rbo}.
Recent learning-based approaches instead exploit tactile feedback.
Luo et al. combine a calibrated tactile simulator, a layout-aware tactile encoder, and object-specific RL experts distilled into a tactile-conditioned diffusion policy, reporting $27\%$ real-world success across $20$ objects~\cite{luo2026blind}.
Lee et al. distill an RL teacher into a transformer policy conditioned on joint positions and uniaxial fingertip forces, reporting high success with a three-fingered gripper mounted on a stationary frame~\cite{lee2026whylook}.

In contrast, our approach shows that a standard hand with rigid contact surfaces can grasp without vision using joint encoders alone.
The key insight is that a short history of joint positions and action commands provides enough information for the policy to reason implicitly about contact, even without tactile, force, or torque measurements.
This minimal sensing setup requires no additional contact sensors and sidesteps the challenge of transferring tactile observations from simulation to hardware.

Proprioception also enables the policy to decide when a grasp is ready for manipulation.
Whereas existing grasp-stability estimators typically learn from explicit outcome labels, such as lift success~\cite{calandra2018more}, our policy learns a scalar transition signal as part of its action space.
The signal communicates grasp readiness to the arm and is shaped by the consequences of initiating manipulation, without direct stability supervision.
Contact reasoning and grasp readiness thus emerge within a single proprioceptive policy.

\section{Methods}
% *think about when writing how it should be applicable to any arm/hand 

% - we should go from high-level picture or idea to low-level details
% - We want to talk about the decomposed architecture (-> the benefit: swapping the arm module)
% - We need to put a big emphasis on the "interface": hand2arm: grasp score -> binary signal -> arm activates different primitives (any primitives); arm2hand: open-or-close-signal -> binary signal -> hand opens/grasps.
% - Key "Learned moduels are grasp score, hand grasp" 
% - Other modules often  can be easily desgined (or even we can use learned primtiives?).

% Intro should talk about the whole figure and whats going on in training and deployment in the high level. Explicity say how you can swap the arm policy during inference without retraining the hand policy. 

% Justify why we need to learn the grasp score/hand action 

% Modualr Architecture (why we need all of these, then details later) justify existence of each module 
% - explain input and output of each module?  
% - grasp score 
% - hand policy
% - reaching arm primitive
% - lifing arm primitive
\begin{figure*}[h]
    \centering
    \includegraphics[width=\linewidth]{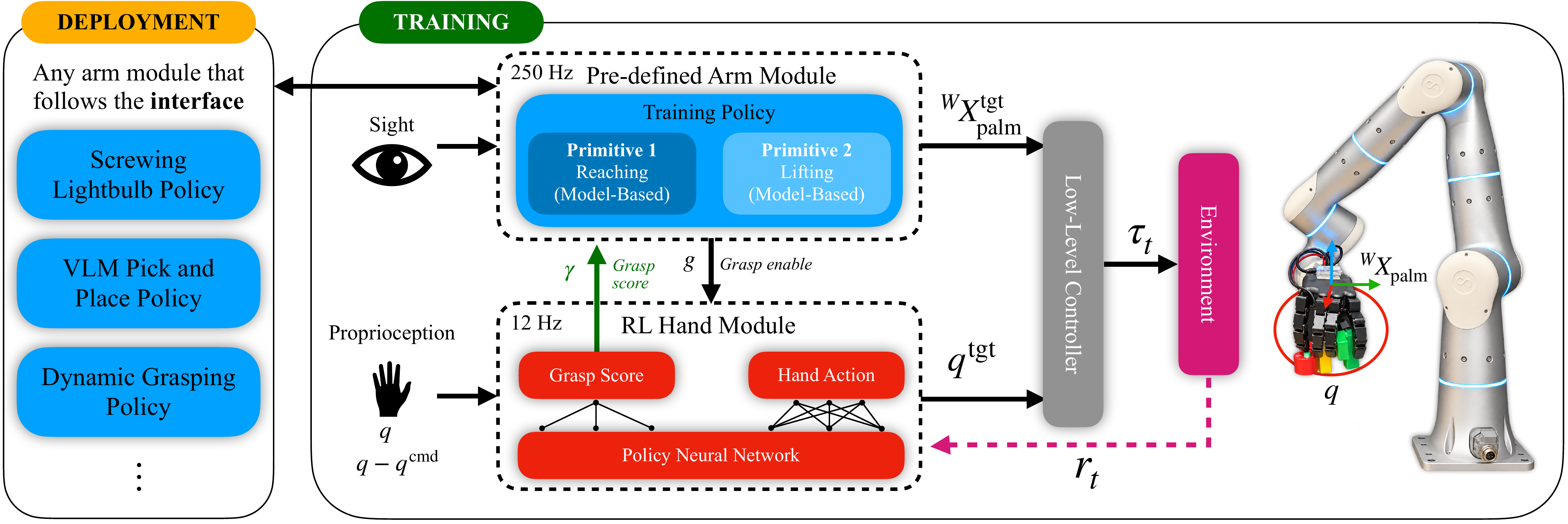}
    \caption{Framework overview. A proprioceptive hand policy coordinates with the arm through an engage command and a learned grasp score, enabling reuse with different arm controllers without retraining.}
    \label{fig:architecture}
    \vspace{-3pt}
\end{figure*}

Our framework separates global arm motion from local grasp control through three modules (\autoref{fig:architecture}). The arm module generates desired palm poses and a binary engage command that activates grasping. The hand module generates finger joint targets and a learned scalar grasp score that signals when to begin post-grasp manipulation, while maintaining and adapting contact as the arm moves the object. The low-level controller converts these outputs into joint commands and limits arm and hand velocities.

We train the hand policy with privileged information in simulation and distill it into a deployable student that observes only a short history of joint positions. Its observations exclude arm state, kinematics, and external perception, although arm motion influences its behavior through contact dynamics. Training uses a fixed two-phase model-based arm controller that reaches and then lifts in response to the grasp score. At deployment, we reuse the frozen hand policy with alternative arm controllers.

\subsection{Learned Hand Module}
\label{sec:hand}

The hand module learns to grasps and stabilizes the object while signaling the grasp score to the arm module. Running at a fixed rate $f_\pi$, it generates hand joint targets and a scalar grasp score. It is the only learned controller in the training setup. Training proceeds in two stages: a privileged teacher learns to grasp using simulated contact and object-state information, and a blind student is distilled from the teacher to reproduce its actions from joint-position history alone. Both policies share the same action space.

\textbf{Policy Action Space.}
The policy outputs raw joint targets $\tilde q_t$ for the hand's $n$ joints along with a grasp score $\graspscore_t \in [0,1]$. To take advantage of a human grasp prior, the neural network's raw joint outputs are moved towards a low-dimensional manifold defined from PCA analysis on a human grasp dataset~\cite{chao2021dexycb}:
% \begin{equation}
%   q^{\mathrm{tgt}}_t = (1-\beta)\,\tilde q_t \;+\; \beta\,\Pi(\tilde q_t),
%   \quad
%   \Pi(q) = (q - \bar q)\,W^{\!\top} W + \bar q,
%   \label{eq:proj}
% \end{equation}
\begin{equation}
q^{\mathrm{tgt}}_t=(1-\beta)\tilde q_t+\beta\Pi(\tilde q_t),\quad
% \Pi(q)=(q-\bar q)W^{\!\top}W+\bar q.
\Pi(q)=W^{\!\top}W(q-\bar q)+\bar q,
\label{eq:proj}
\end{equation}
where $W \in \mathbb{R}^{k \times n}$, $k \ll n$, contains the retained PCA directions, and $\bar q$ is the mean posture. $q^{\mathrm{tgt}}_t$ is the final joint position target of the hand module, which is then passed to the low-level velocity controller. We construct this synergy basis by fitting PCA to hand postures retargeted from human grasp demonstrations onto the robot hand's kinematics. The blending weight $\beta \in [0,1]$ controls the strength of the prior: intermediate values encourage coordinated finger motion while allowing departures from the manifold, whereas $\beta=1$ constrains targets to the manifold.

\subsubsection{Privileged Teacher}

The teacher is trained with RL on the task reward using privileged observations available in simulation:
\begin{equation}
  o^{\mathrm{teacher}}_t = \big[\, (c^{\mathrm{full}}_t)^{\!\top},\;\; (d^{\mathrm{palm}}_t)^{\!\top},\;\;
  q_t^{\top}\,\big]^{\top} \in \mathbb{R}^{L+3+n},
  \label{eq:obs-teacher}
\end{equation}

where $c^{\mathrm{full}}_t \in \{0,1\}^{L}$ indicates binary contact on the hand's $L$ links, $d^{\mathrm{palm}}_t \in \mathbb{R}^{3}$ is the object's position in the palm frame, and $q_t$ denotes the joint angles of the hand. Contact indicators identify which links touch the environment, while the relative object position guides grasp acquisition. These privileged inputs facilitate learning before distillation removes their use at deployment. An asymmetric critic additionally observes the object's world position and velocity.

\subsubsection{Distillation to a Blind Student}

We distill the teacher into a student that predicts the teacher's full action, including joint targets and the grasp score, from a short history of measured joint angles and the previous position command:
% \begin{equation}
%   o_t = \big[\,q_{t-H},\; \dots,\; q_{t-1},\;
%   q_{t-H} - q^{\mathrm{cmd}}_{t-H}, \; \dots, \; q_{t-1} - q^{\mathrm{cmd}}_{t-1}\,\big]
%   % q^{\mathrm{tgt}}_{t-H+1}, \; \dots, \; q^{\mathrm{tgt}}_{t}\,\big]
%   \;\in\; \mathbb{R}^{(2H)n},
%   \label{eq:obs}
% \end{equation}
% where $H$ is the number of joint-position observations in the history and $q^{\mathrm{cmd}}_{t-1}$ is the commanded configuration issued at the previous policy step for the low-level torque controller. The student therefore uses joint-encoder measurements and its own previous command, without tactile, force, torque, or object-state measurements.
% \begin{equation}
% o^{\mathrm{student}}_t = \big[
% q_{t-H}, \dots, q_{t-1},
% \Delta q_{t-H}, \dots, \Delta q_{t-1}
% \big] \in \mathbb{R}^{2H \times n},
% \label{eq:obs}
% \end{equation}
\begin{equation}
o^{\mathrm{student}}_t = \big[
q_{t-H}^{\top}, \dots, q_{t-1}^{\top},
\Delta q_{t-H}^{\top}, \dots, \Delta q_{t-1}^{\top}
\big]^\top \in \mathbb{R}^{2Hn},
\label{eq:obs}
\end{equation}
% \tyl{$\mathbb{R}^{2H \times n}$?}
where $H$ is the history length, $n$ is the number of hand joints, and $\Delta q_k = q_k - q_k^{\mathrm{cmd}}$ is the difference between the measured and commanded joint configurations at step $k$.
The student therefore uses joint-encoder measurements and its own previous command, without tactile, force, torque, or object-state measurements.

Together, motion history and the previous command provide cues for implicit contact reasoning. The command indicates the intended finger configuration, while the measured trajectory reveals how the hand responds. Contact can slow or arrest a finger's motion, creating a persistent discrepancy between its target and measured position. Through teacher supervision, the student learns to interpret these cues over time and adjust its grasp accordingly. This observation design separates local grasp control from global arm control, enabling the use of different arm controllers during inference.

\subsection{Implicit Learned Grasp Score}
\label{sec:score}

% The grasp score connects local grasp control to the arm's decision to begin manipulation. Alongside the joint targets, the hand policy outputs a scalar $s_t = \sigma(a^{(0)}_t) \in [0,1]$, which controls the transition between reaching and lifting during training. At deployment, the same signal can trigger subsequent manipulation motions, such as sliding or pulling.
The grasp score connects local grasp control to the arm's decision to begin manipulation. Alongside the joint targets, the hand policy outputs a scalar $\graspscore_t \in [0,1]$, which controls the transition between reaching and lifting during training. At deployment, the same signal can trigger subsequent manipulation motions, such as sliding or pulling.

% The training-time arm controller starts in the reaching phase. At each step, it selects reaching when $s_t < s_{\mathrm{lo}}$, selects lifting when $s_t > s_{\mathrm{hi}}$, and retains its previous phase otherwise, where $s_{\mathrm{hi}} > s_{\mathrm{lo}}$. This hysteresis prevents small fluctuations around a single threshold from repeatedly switching the controller. The transition is reversible: if the score falls below $s_{\mathrm{lo}}$ during lifting, the arm returns to reaching while the hand continues attempting to establish a grasp. This rule permits recovery but does not guarantee that the learned score detects every deterioration in grasp quality. The reaching and lifting controllers are described in the next section.

During training, the grasp score controls the arm's transition
between reaching and lifting through a fixed two-state finite-state
machine (FSM). The arm controller starts in reaching. Given predefined
thresholds $\graspscore_{\mathrm{lo}} <
\graspscore_{\mathrm{hi}}$, it selects reaching when
$\graspscore_t < \graspscore_{\mathrm{lo}}$, lifting when
$\graspscore_t > \graspscore_{\mathrm{hi}}$, and retains its
previous phase otherwise. The transition is
reversible: if the score falls below
$\graspscore_{\mathrm{lo}}$ during lifting, the arm returns to
reaching while the hand continues attempting to grasp.

% The training-time arm controller starts in the reaching phase. At each step, it selects reaching when $\graspscore_t < \graspscore_{\mathrm{lo}}$, selects lifting when $\graspscore_t > \graspscore_{\mathrm{hi}}$, and retains its previous phase otherwise, where $\graspscore_{\mathrm{hi}} > \graspscore_{\mathrm{lo}}$. This hysteresis prevents small fluctuations around a single threshold from repeatedly switching the controller. The transition is reversible: if the score falls below $s_{\mathrm{lo}}$ during lifting, the arm returns to reaching while the hand continues attempting to grasp. This rule permits recovery but does not guarantee that the learned score detects every deterioration in grasp quality.

The teacher learns this score as an action dimension through RL, without explicit grasp-stability labels or an auxiliary classification loss. Its value is shaped by the consequences of switching arm phases under the task reward. Initiating a lift before securing the object can reduce the height and hold rewards, while delaying a viable lift postpones those rewards. Training therefore encourages a transition signal that reflects readiness for the lifting behavior encountered during training.

During distillation, the student learns to reproduce the teacher's score through action supervision, using joint-position history and the previous position-target command. Thus, the teacher learns the transition decision from task consequences, and the student learns to approximate that decision without the teacher's privileged contact and object-state observations. Therefore, we interpret $\graspscore_t$ as a proxy to grasp stability.

\subsection{Arm Module}
\label{sec:arm}

The arm module coordinates reaching and manipulation through a compact interface with the hand. It receives the grasp score $\graspscore_t$ and outputs a desired palm pose ${}^WX^{\mathrm{palm}}_{\mathrm{des}} \in SE(3)$ for the low-level controller, together with a binary engage signal $g_t$. When $g_t = 0$, the hand is commanded to a fixed, splayed pre-grasp posture that overrides $q^{\mathrm{tgt}}$; when $g_t = 1$, the learned command in Eq.~\eqref{eq:proj} is applied. The engage signal activates grasp control, while the grasp score signals grasp stability for post-grasp manipulation. Neither module requires access to the other's internal state: the hand does not observe the arm's kinematics, plan, or perception, and the arm uses the grasp 
score to coordinate its motion. 

\textbf{Training-time instantiation.}
During training, a fixed two-phase model-based controller gives the grasp score a consistent consequence. In the reach phase, the arm servos the palm toward a fixed pose offset ${}^{\mathrm{palm}}\Delta X \in SE(3)$ in the palm frame from the object. When the palm is within a fixed constant distance of the estimated object position, it signals $g_t = 1$ to start grasping. In the lift phase, it moves toward a fixed pose at a height $h_{\mathrm{lift}}$ above the home configuration. The grasp score selects the phase, allowing the hand policy to learn when to initiate or re-try a lift.

\textbf{Deployment-time instantiation.}
This interface allows the frozen hand policy to be paired with compatible arm controllers, including scripted trajectories, motion planners, reactive controllers, teleoperation, and vision--language policies. Compatibility requires both the specified signal exchange and arm motions under which the hand can establish and maintain contact. We evaluate several controller substitutions without retraining the hand policy, including a vision--language arm module for language-conditioned pick-and-place in clutter, and a controller to screw in a lightbulb. In each of these cases, the arm commands $g_t = 1$ to begin grasping based on a distance threshold, and commands $g_t = 0$ when the hand should let go, such as when the object is near the target position or the hand needs to let go of the lightbulb for another rotation.

\subsection{Low-Level Velocity Controller}
\label{sec:lowlevel}

The low-level controller converts the arm's desired palm pose and the hand's joint targets into actuator commands. This separation exposes tracking gains and velocity limits for hardware tuning while keeping the hand policy fixed.

\textbf{Arm.}
Following~\cite{lee2025hierarchical}, we convert the palm pose error into a bounded desired twist that decreases as the palm approaches its target. During training, damped least-squares inverse kinematics (IK) maps this twist to joint velocities, supporting efficient batched simulation. 
On hardware, we instead solve a quadratic program (QP) that tracks the desired twist subject to joint position, velocity, and linearized collision-avoidance constraints.

\textbf{Hand.} We construct a velocity command $\dot q^{\mathrm{cmd}}$ from the policy output $q^{\mathrm{tgt}}$ using:
\begin{equation}
  \dot q_t^{\mathrm{cmd}} =
  \begin{cases}
    v_{\max}\,
    \dfrac{q_t^{\mathrm{tgt}} - q_t^{\mathrm{cmd}}}{d},
    & \text{if } |q_t^{\mathrm{tgt}} - q_t^{\mathrm{cmd}}|
      \le d, \\[6pt]
    v_{\max}\,
    \mathrm{sgn}\!\left(q_t^{\mathrm{tgt}} - q_t^{\mathrm{cmd}}\right),
    & \text{otherwise},
  \end{cases}
  \label{eq:handvf}
\end{equation}
% The hand controller gradually advances an internal position command $q^{\mathrm{cmd}}$ toward the policy target $q^{\mathrm{tgt}}$ using a velocity field defined by a maximum velocity $v_{\max}$ and a threshold distance $d_{\mathrm{slow}}$ which it starts linearly decaying the velocity near the target also following \cite{lee2025hierarchical}.
% \begin{figure}[!t]
%     \centering
%     \includegraphics[width=1.0\linewidth]{tabs/figures/fig_action_chain.pdf}
%     \caption{
%     CAPTION 
%     \comment{Change $q^{lin}$ to $\tilde q$, Eq. (2) to Eq. (1), Eq. (12) to Eq. (4)}
%     }
%     \label{fig:cmd}
%     \vspace{-2mm}
% \end{figure}
where $v_{\max}$ limits the command speed and $d>0$ sets the
distance over which the command slows near the target. We update
the command as
$q_t^{\mathrm{cmd}} = q_{t-1}^{\mathrm{cmd}}
+ \dot q_{t-1}^{\mathrm{cmd}} \Delta t$,
where $\Delta t$ is the low-level control period. This update
uses the previous command rather than the measured joint position.
When contact stops a finger, $q^{\mathrm{cmd}}$ can therefore
continue toward $q^{\mathrm{tgt}}$, maintaining a tracking error
that produces joint effort through the servo.
% This command structure provides temporal cues for the student policy. Joint-position history and the previous policy target reveal how actual motion differs from intended motion, supporting implicit contact reasoning. The student does not directly observe the internal servo command, and these cues do not uniquely identify contact.

For both the hand and arm, the resulting commands are tracked using PD feedback with gravity compensation.
% \begin{equation}
%   \tau = K_{p}\!\left(q^{\mathrm{cmd}} - q\right)
%        + K_{d}\!\left(\dot q^{\mathrm{cmd}} - \dot q\right)
%        + \tau_{g}(q),
%   \label{eq:pdtorque}
% \end{equation}
\begin{equation}
\tau = K_p(q_{\mathrm{all}}^{\mathrm{cmd}} - q_{\mathrm{all}})
     + K_d(\dot q_{\mathrm{all}}^{\mathrm{cmd}} - \dot q_{\mathrm{all}})
     + \tau_g,
\label{eq:pdtorque}
\end{equation}
where $q_{\mathrm{all}} = [q_{\mathrm{arm}}^\top, q^\top]^\top$
is the combined arm--hand joint configuration, $K_{p}$ and $K_{d}$ are diagonal stiffness and damping gain matrices and
$\tau_{g}(q)$ is the gravity compensation torque.

\begin{table}[t]
\caption{Task reward terms. }
\label{tab:rewards}
\centering
% \small
\setlength{\tabcolsep}{4pt}
\renewcommand{\arraystretch}{1.2}
% \begin{tabular}{@{}llp{2.6cm}@{}}
\begin{tabular}{lll}
\toprule
\textbf{Term} & \textbf{Definition} & \textbf{Purpose} \\
\midrule
Coverage & $\sum_i w_i \, \mathbf{1}\!\left[\,\ell_i\,\right]$
         % & Establish and maintain inner-hand contact \\
         & Inner-hand contact \\
Height   & $\mathrm{clip}\!\left(z_t-z_{\mathrm{rest}},\,0,\,z_{\max}\right)$
         % & Encourage object lifting \\
         & Object lifting \\
Hold     & $\mathbf{1}\!\left[\,\|p^{\mathrm{obj}}_t-p^{\mathrm{goal}}\|<d_{\mathrm{tol}}\,\right]$
         % & Encourage residence near the goal \\
         & Stable hold \\
\bottomrule
\end{tabular}
\par\vspace{3pt}
\parbox{\linewidth}{\footnotesize
$\ell_i$ indicates inner-surface contact on link $i$, weighted by $w_i$. $z_{\mathrm{rest}}$ is the object's settled height at episode start, $z_{\max}$ caps the height reward, and $p^{\mathrm{goal}}$ is a fixed goal above the table with tolerance $d_{\mathrm{tol}}$.
}
\end{table}

{\setlength{\textfloatsep}{6pt}
\begin{table*}[!t]
  \centering
  \caption{Comparison against baselines. The end-to-end policy leads on static
  grasping, but collapses on dynamic objects and covers only part of the
  workspace; our method is within a few points of it on static success while
  retaining both.}
  \label{tab:main}
  \begin{tabular*}{\textwidth}{@{\extracolsep{\fill}}lccccccc@{}}
    \toprule
    & \multicolumn{2}{c}{Objects grasped (30\,s)} & Per-attempt & Dynamic & Object pose & Arm  & Re-grasp\\
    \noalign{\vskip -1.5pt}
    \cmidrule(lr){2-3}
    \noalign{\vskip -1.9pt}
    Method & YCB ($n{=}78$) & GraspXL ($n{=}3028$) & success & success & generalization & agnostic & ability\\
    \midrule
    Scripted              & 31\%          & 39\%          & 39\%          & 15\%          & 30\% & \yes & \no \\ 
    RobustDexGrasp \cite{zhang2025robustdexgrasp}$^\dagger$          & \textbf{98\%} & \textbf{97\%} & \textbf{85\%} & 8\%           & 15\%  & \no  & \yes \\
    \midrule
    Ours (no grasp score) & 73\%          & 82\%          & 82\%          & 75\%          & 70\% & \yes & \no \\
    \textbf{Ours}         & 96\%          & 95\%          & \textbf{85\%} & \textbf{92\%} & \textbf{90}\% & \yes & \yes \\
    \bottomrule
  \end{tabular*}
  % \vspace{2pt}
  \raggedright
  \footnotesize $^\dagger$Their teacher policy was trained in our blind setting with our reward formulation.
\end{table*}

\subsection{Policy Training}
\label{sec:training}

% \textbf{Environment.}
% We train in a GPU-parallel rigid-body simulator with multiple environments stepped simultaneously. Objects are drawn from a mesh dataset and scaled uniformly to vary their size. At each episode, we resample object mass, planar position, and yaw, and perturb the palm offset ${}^{\mathrm{palm}}\Delta X$ at grasp initiation within bounded position and orientation ranges. This randomization encourages robustness to object variation and palm pose.

\textbf{Rewards.}
The task reward $r_t$ is a weighted sum of contact coverage, height, and hold terms, as listed in \autoref{tab:rewards}. The rewards are multiplied by the control period to account for the duration of each step. \textit{Coverage} rewards establishing and maintaining inner-hand contact. \textit{Height} provides a dense incentive to raise the object, which helps with exploration early on. \textit{Hold} rewards each step spent near an elevated goal. Repeated hold rewards encourage sustained grasps near the goal and more stable grasps. We have no rewards for action rate or energy penalties, as our velocity-limiting low-level controller ensures smooth and controlled motion. 

\textbf{Teacher.}
We train the privileged teacher using PPO~\cite{schulman2017ppo}. The actor and asymmetric critic each consist of a single-layer GRU followed by an MLP. Recurrence allows the teacher to integrate contact and motion information over time when selecting joint targets and the grasp score.

\textbf{Student.}
We distill the frozen teacher using DAgger~\cite{ross2011dagger}. The student acts in the environment, and the teacher provides action targets at the states reached by the student. Training on these trajectories exposes the student to the consequences of its own actions. The student minimizes a mean-squared error over the full action vector, including the grasp score, using joint-position history and the previous position-target command as inputs. It thereby learns both grasp control and the teacher's transition signal without privileged contact or object-state observations.

\section{Experiments}

% Experimental setup 
% - hardware
% - training details (sim used, frequencies of low and high level, how many objects trained on)

% Baselines
% Cols: Grasp success rate, sensitivity to pose error, sensitivity to table height 
% - naive baseline (go to object, close and lift) 
% - vision based end to end (arm and hand) RL (assume know full object state, or like in RDG)
% - model based grasping approach (plan grasp pose from geometry) 
% - our method 

% would be cool to show our method is better at dealing with pose error than the naive baseline (where you think the object is somewhere else) 

% Ablations
% Cols: Grasp success rate, grasp score spearman correlation, sensitivity to Kp/V\_max  
% Teacher policy 
% PCA (0, 0.5, 1.0)
% MLP vs RNN
% proprioception observation (q\_targ + q, q\_cmd + q, q)

% Using different arm modules 
% - Standard used during training
% - VLM pick and place 
%include frames from video 
% - Screwing lightbulb 
% include figure of three frames of lightbulb screwing 

% Grasp Score Analysis 
% - Show spearman correlation and graph with force closure metric versus grasp score 

\subsection{Experimental Setup}
\label{sec:exp-setup}

\textbf{Hardware.}
We instantiate the framework on a 7-DoF Flexiv Rizon 4 torque-controlled arm equipped with a
20-DoF Robotis HX5-D20-MRT five-finger anthropomorphic hand. The hand, arm, and low-level velocity controller are run as separate asynchronous processes. The
low-level controller runs at $1$\,kHz, the arm module at $200$\,Hz, and the hand policy
at the same $11.9$. 

\textbf{Simulation and training.}
Teacher and student are trained in MuJoCo Warp~\cite{mujocowarp}, with 4096 worlds stepped in parallel on a single GPU. The physics
timestep is $4$\,ms ($250$\,Hz) and the policy acts every $21$ physics steps, giving a
control period of $84$\,ms ($11.9$\,Hz), and each episode lasts $10$\,s.

Training uses the full YCB object set~\cite{calli2015ycb}, instantiated at several
uniform scale factors so that object identity and size vary across the parallel
population. We apply domain randomization to the object's mass, pose, and scale, to the
palm offset ${}^{\mathrm{palm}}\Delta X$, to external velocity disturbances applied to the
object, and to the velocity limit and servo gains that the hand policy
acts through.

The teacher is a single-layer GRU with $256$ hidden units followed by a
$(256,128,64)$ MLP with ELU activations, observing the $44$-dimensional privileged set
of Eq.~\eqref{eq:obs-teacher} and optimized with PPO. The student
is a feed-forward $(256,128,64)$ MLP over the $200$-dimensional stack of $H=5$ joint
position and error frames of Eq.~\eqref{eq:obs}. Both teacher and student emit the
same $21$-dimensional action, one grasp score and $20$ joint targets, with the
posture-manifold projection at $\beta = 0.5$. Teacher and student training required together approximately 4 
GPU-hours on a single Nvidia L40S GPU.

\textbf{Evaluation protocol.}
To mimic a real-world grasping scenario, we give each method a 30-second window to grasp the object, where it has the opportunity to re-grasp if it detects failure. We count a success if the object is held within 10 cm of the target position above the ground for 2 seconds. Even though the policies are trained using a static object pose estimate, we want to see if they can generalize to when the commanded pose changes mid-grasp. To measure this we introduce a dynamic success rate metric. Using a set of 10 approximately spherical objects from the YCB dataset, we spawn them at a random location within the trained grasp area, and once the method starts running the object is moved to another random spot in the trained grasp area, and the grasp success rate is measured. This decouples this metric from our object pose generalization metric, where we spawn the objects outside the trained grasp area to see how well each method generalizes to unseen grasp locations. 

\begin{figure}[!t]
    \centering
    \includegraphics[width=1.0\linewidth]{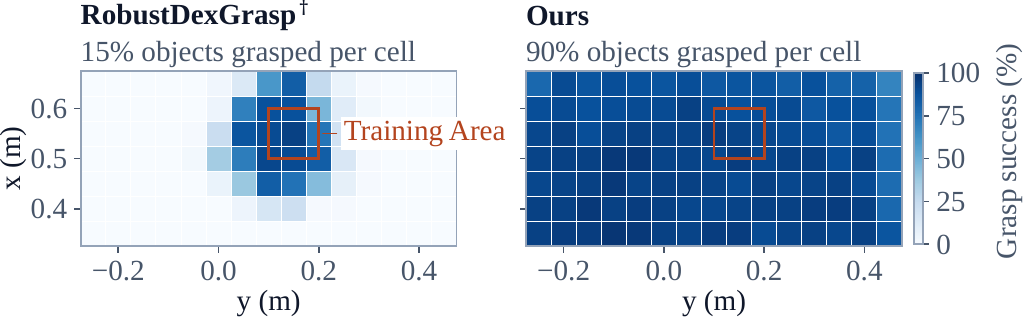}
    \caption{
    Grasp success as a function of where the object rests on the table ($x$, $y$ in
metres from the robot base), for our method and the end-to-end baseline. The arm starts
each episode at its home pose, and both policies saw the object only inside the training
spawn box (orange).
    }
    \label{fig:workspace}
    % \vspace{-2mm}
\end{figure}

\begin{figure*}[!t]
    \centering
    \includegraphics[width=1.0\linewidth]{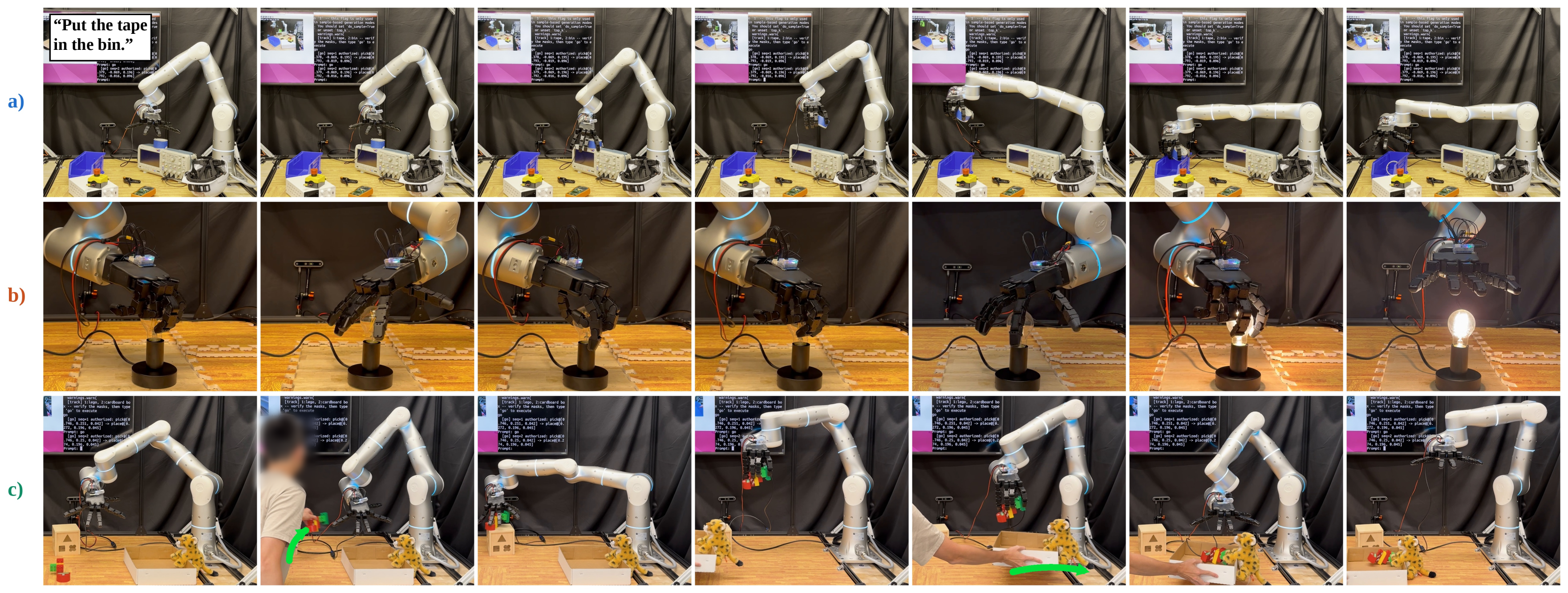}
    \caption{
    a) The VLM pick-and-place policy running on the arm moving tape from the top of the oscilloscope to the bin. b) The policy to screw in a lightbulb. c) Dynamic grasping where the object location and target location are both changed while the policy is running. 
    }
    \label{fig:arm_modules}
    % \vspace{-2mm}
\end{figure*}

\subsection{Baselines}
\label{sec:exp-baselines}
We compare against three alternative approaches to solve the problem of blind grasping.

\emph{Scripted closure} is where the arm moves to its
believed object position, the fingers close at a fixed rate until the actuators
saturate, and the arm lifts on a timer. The open and close poses are manually tuned.

\emph{Vision-based end-to-end RL} trains a single policy over the arm and hand jointly from object point cloud data. We adapt the teacher policy
of RobustDexGrasp~\cite{zhang2025robustdexgrasp} to our setting.
In the static evaluation, the policy receives the initial
object point cloud throughout the episode. In the dynamic evaluation,
the point cloud is updated after the object moves, allowing the policy
to observe the new object location. We evaluate the trained teacher
directly, without distillation.

\emph{Our method without grasp-score feedback} uses the same
trained hand policy but ignores the grasp score. The arm
begins lifting 5\,s after grasping is activated, replacing
the learned transition with a fixed timer, as in~\cite{agarwal2023dexterous}.

\subsection{Grasping Performance}
% \subsection{Grasp Success Rate}
\label{sec:exp-grasp-success-rate}
\autoref{tab:main} shows the grasp success rate. The end-to-end policy performs strongly in the static setting, where
the arm and hand are optimized jointly for grasping. Despite using a
separately designed arm controller with a fixed grasp approach, our
method achieves comparable grasping performance while retaining the
modularity of the arm and hand. The difference becomes more apparent when the object location changes
during execution or when grasps are attempted outside the training
region. In these settings, the end-to-end baseline degrades
substantially even when provided with updated point-cloud observations.
This degradation is consistent with distribution shift: the end-to-end baseline
was trained on static object locations, and the resulting corrective
arm-hand motions are outside its training distribution. In contrast,
our hand policy remains unchanged while the arm controller can update
the palm target online to react to the new object location. A similar trend is observed in~\autoref{fig:workspace}. The end-to-end baseline performs well near the training region but degrades as the object location moves outside the distribution seen during training, whereas our factorized approach maintains successful grasping over a broader workspace.

\begin{figure}[!t]
    \centering
    \includegraphics[width=0.7\linewidth]{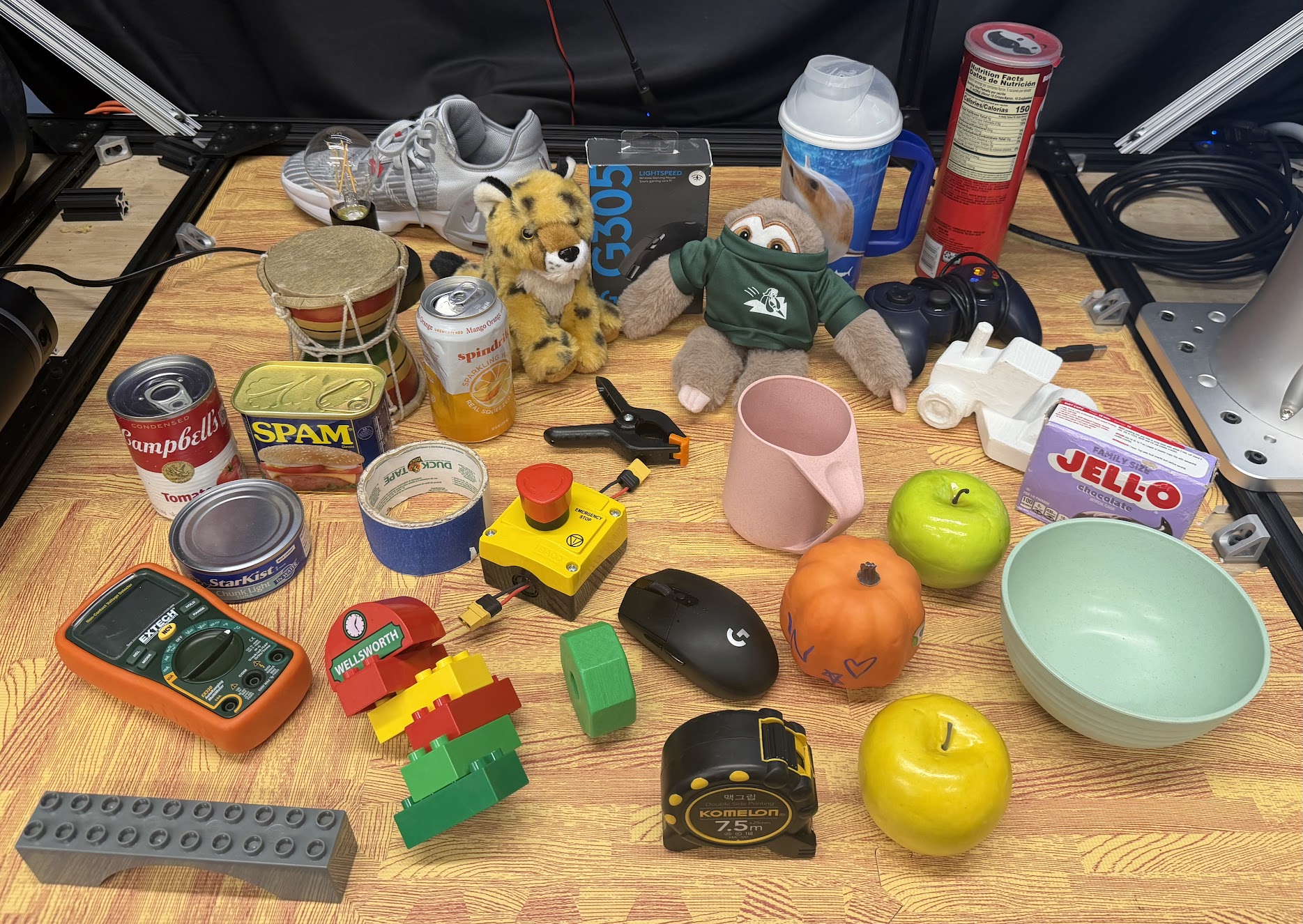}
    \caption{
    29 real-world objects that our method grasped during testing.
    }
    \label{fig:realworld-objects}
    % \vspace{-2mm}
\end{figure}

We additionally test the policy on 29 real-world objects with
different shapes, sizes, and materials
(see \autoref{fig:grid} and \autoref{fig:realworld-objects}). This experiment is intended as a
qualitative test of object diversity rather than a success-rate
benchmark, and we therefore do not report repeated trials for each
object. The same frozen hand policy successfully grasped all 29
objects, without providing object geometry to the hand
policy. We refer the reader to the accompanying multimedia video for qualitative results.
% Firstly, we expect the highest percentage of objects solved for the end-to-end trained policy, as the arm and hand are optimized jointly as opposed to using a fixed palm pose for every object. However, even with the fixed palm pose used in our method, our hand policy can still achieve comparable success rates. Because the end-to-end method also controls the arm, it drastically fails in the dynamic object scenario, which was never encountered during training. Furthermore, it shows poor generalization to object pose as shown in \autoref{fig:workspace}, only being able to grasp in a small portion of the robot workspace. Being arm agnostic, our method has many advantages, further outlined below, including but not limited to allowing the same trained hand policy to work with different robot arms (cross-embodiment), different arm mounting poses, and different arm controllers. If we apply our method but omit the grasp score and instead lift based on time as in \cite{agarwal2023dexterous}, then we get the advantages of being arm agnostic but limit our grasping capability, as we aren't able to detect failed grasps and re-grasp the object. By using a hierarchical approach for the arm and hand but still allowing two-way communication as outlined by our interface with the grasp score, we can achieve a highly versatile policy that is also robust to grasp failures.  

\subsection{Grasp Score Analysis}
\label{sec:exp-score}

To understand what the learned grasp score captures, we compare it
against the Ferrari-Canny $\epsilon$ metric~\cite{ferrari1992planning},
a classical measure of grasp quality. The metric evaluates the ability
of a set of contacts to resist external disturbances. A larger
$\epsilon$ indicates that the grasp can resist a larger worst-case
disturbance wrench across directions. We compute $\epsilon$ from the ground-truth
contacts available in simulation. These contacts are never observed by
the deployed policy, and $\epsilon$ is not used in the reward or as a
training target. As a result, agreement between the two signals is not explicitly
optimized.

\autoref{fig:grasp_score} shows the learned grasp score and Ferrari-Canny averaged across successful and failed grasp attempts. We align each trajectory at the lift transition, defined as the first time
$\graspscore_t$ crosses the switching threshold. For successful grasps, both
signals remain high after the lift begins. For failed grasps, both decrease
after the transition as the grasp deteriorates.  The similar behavior of the two signals suggests that the learned grasp score captures changes in grasp quality,
despite having no direct supervision from the contact information used to
compute $\epsilon$.

We quantify this agreement using the Spearman rank correlation $\rho$~\cite{spearman1904proof} between $\graspscore_t$ and $\epsilon$ within each episode, averaged over episodes. Spearman correlation measures the monotonic association between two variables based on their ranks, making it appropriate here because the learned score is not calibrated to the scale of the physically derived $\epsilon$ metric and their relationship need not be linear. The score agrees strongly on both cohorts, with $\bar\rho = +0.78$ on successes ($n=14346$) and $\bar\rho = +0.61$ on failures ($n=751$), yielding $\bar\rho = +0.77$ over all episodes.

\begin{figure}[!t]
    \centering
    \includegraphics[width=1.0\linewidth]{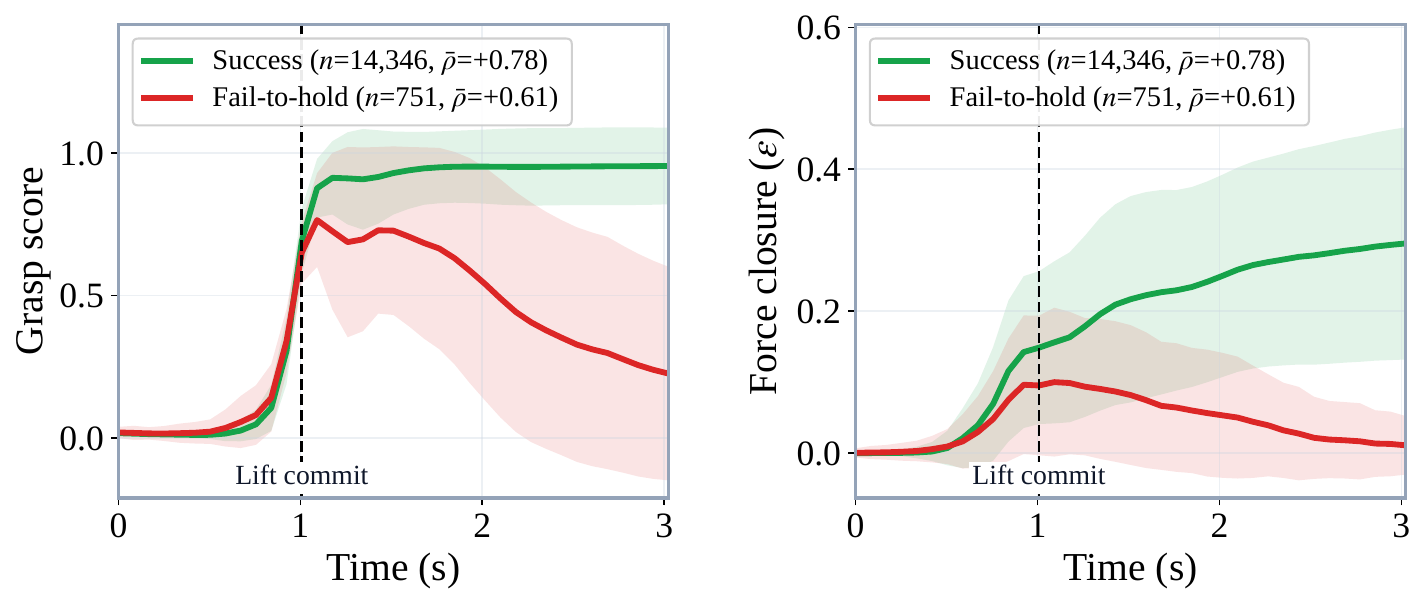}
    \caption{
    Learned grasp score (left) and Ferrari--Canny $\epsilon$ (right) averaged
across grasp attempts. The score is never supervised and never observes
the contacts $\epsilon$ is computed from, yet the two separate at the same instant and
decay together on failures; $\bar\rho$ is the within-episode Spearman correlation
between them, averaged over episodes.
    }
    \label{fig:grasp_score}
    % \vspace{-2mm}
\end{figure}

\subsection{Composing with Different Arm Modules}
\label{sec:exp-arms}

We demonstrate hand-policy reuse by pairing the same frozen hand policy
with arm behaviors not seen during training
(\autoref{fig:arm_modules}). 
A vision--language controller uses
RexOmni~\cite{jiang2026detect} to select diverse objects and placement
locations from natural-language instructions in cluttered scenes, with
depth used to recover 3D palm targets. A reactive controller uses
SAM2~\cite{ravi2025sam} to track the object online and continuously
update the palm target as the object moves. We also use a scripted
lightbulb-screwing routine, which introduces sustained wrist rotations
and arm motions that are substantially different from the vertical lift
used during training. In all cases, the hand policy remains unchanged
and uses the same bidirectional interface. The arm engages or releases
the grasp, while the grasp score indicates when the hand is ready for
subsequent manipulation. For example, the lightbulb routine begins
rotation only after the grasp score crosses the threshold and releases
the object after each rotation so the sequence can be repeated.
% The 29 objects used in the hardware experiments are shown
% in~\autoref{fig:realworld-objects}.

We further test whether the same hand policy can be reused across arm
configurations and platforms. As shown in \autoref{fig:mounts}, the
policy successfully grasps with arm mounting configurations not seen
during training and with different robot arms. These experiments change
the upstream arm kinematics, mounting geometry, and control behavior
without retraining the hand policy.

% First, a vision-language module performs language-conditioned pick-and-place in a cluttered scene, selecting targets and issuing palm setpoints while the hand remains blind. Next we show a reactive grasping controller using SAM2~\cite{ravi2025sam} to track the object and target in real time and reactively move the palm to adjust. Finally, a scripted manipulation routine screws a lightbulb into a socket, a task that requires the grasp to persist through sustained wrist rotation rather than a vertical lift. Importantly, all modules take advantage of the two way communication of our interface, for example the lightbulb is only rotated once the grasp score is reported above a certain threshold, and then after rotation the arm commands the hand to stop grasping so the routine can be repeated.

In simulation, we further show that rather than using the pseudo-inverse velocity controller for the robot, which was used during training, at deployment we can solve a QP adapted from \cite{lee2026reactive} which considers link-to-collision-object distances as well as joint limits to perform online collision avoidance. We demonstrate this in \autoref{fig:obstalces} where the arm can successfully grasp even with three obstacles.

\begin{figure}[!t]
    \centering
    \includegraphics[width=1.0\linewidth]{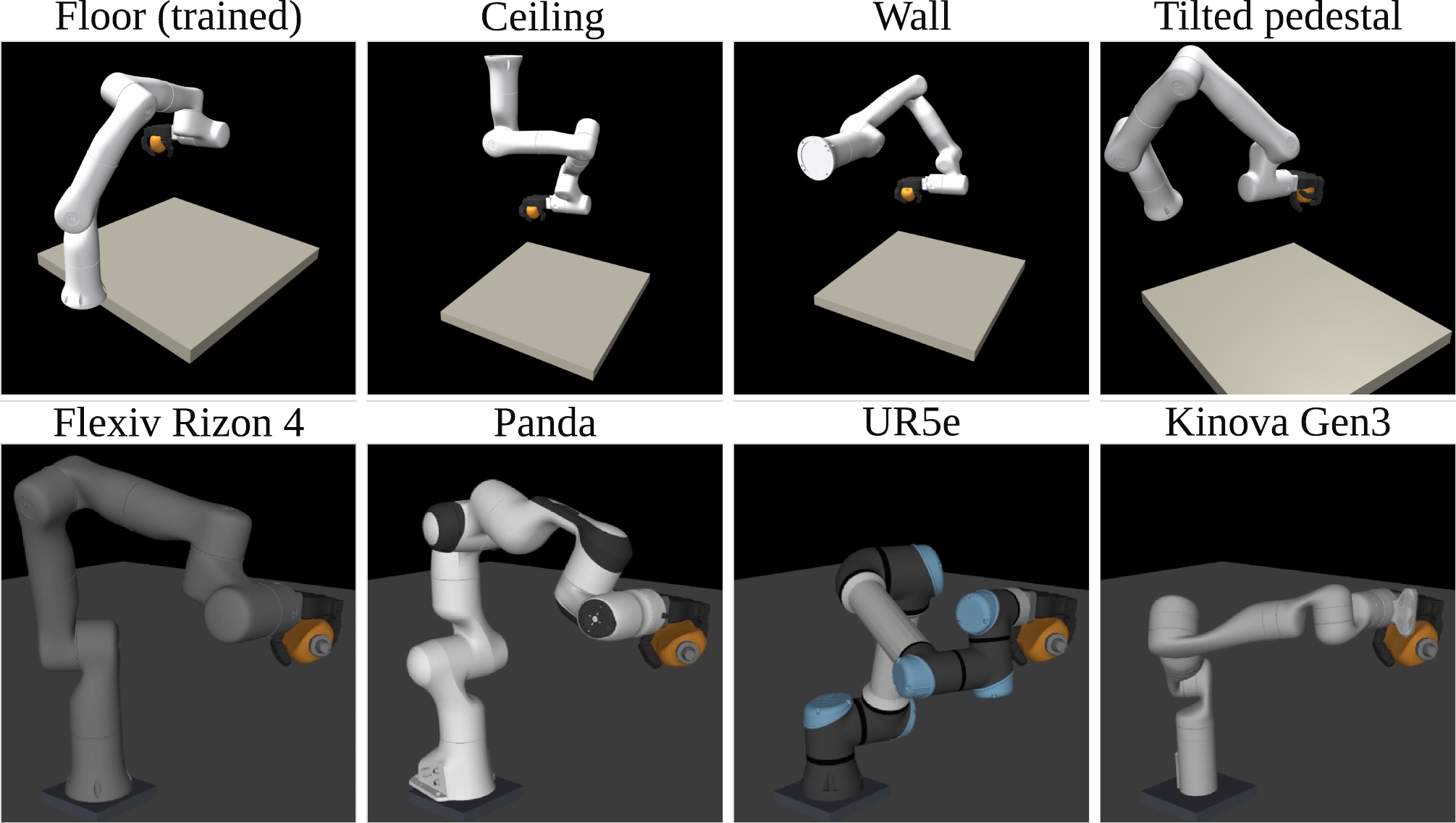}
    \caption{
    (Top) Arm successfully grasping with different mounting positions not seen during training. 
    (Bottom) Cross-embodiment across three different robot arms not seen during training. 
    }
    \label{fig:mounts}
    % \vspace{-2mm}
\end{figure}

\begin{figure}[!t]
    \centering
    \includegraphics[width=1.0\linewidth]{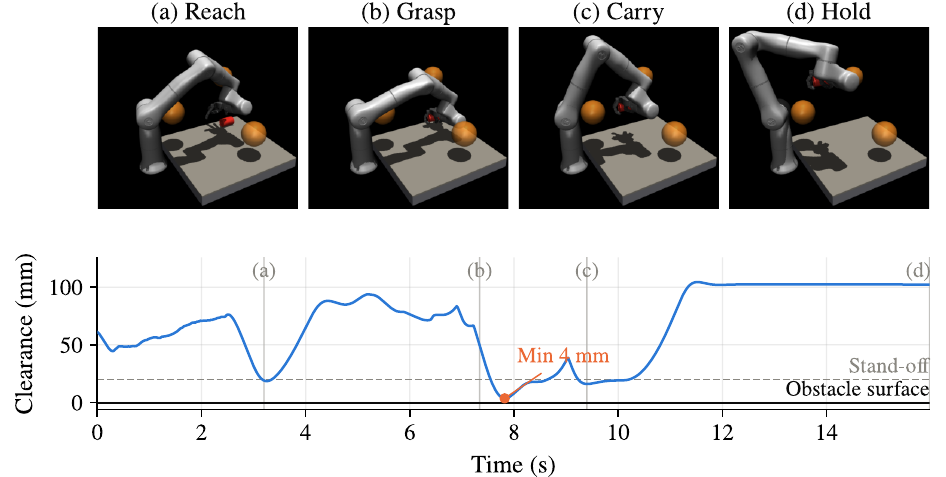}
    \caption{
    At deployment, QP is solved to compute joint velocities that satisfy the task space velocity command and also avoid collisions with obstacles in real-time. 
    }
    \label{fig:obstalces}
    % \vspace{-4mm}
\end{figure}

% \subsection{Arm-Agnostic Behavior}

\subsection{Ablations}
\label{sec:exp-ablations}

\autoref{tab:ablations} isolates the contribution of each design decision. First, we compare the PCA projection constant $\beta$ to adjust how much we push the policy output towards the PCA manifold. Second, we try three different observations for the student policy. Finally, we show how our method generalizes to different $k_p$ and $v_{\max}$. For each variant report the success rate on the YCB dataset, the time until success, and the Spearman correlation with the force closure metric.

{\setlength{\textfloatsep}{6pt}
\begin{table}[t]
  \centering
  \caption{Ablation results. Each block varies one design choice. \textbf{bold} indicates the design choice of the proposed method.}
  \label{tab:ablations}
  \begin{tabular*}{\columnwidth}{@{\extracolsep{\fill}}lccc@{}}
    \toprule
    & Success & Time to hold (s) & $\bar\rho$ \\
    \midrule
    \multicolumn{4}{@{}l}{\textit{Action space (PCA)}} \\
    $\beta = 0$             & 91\%          & 3.8          & 0.72          \\
    $\beta = 0.5$  & \textbf{96\%} & \textbf{3.9} & \textbf{0.77} \\
    $\beta = 1$         & 28\%          & 3.9          & 0.70          \\
    \midrule
    \multicolumn{4}{@{}l}{\textit{Student observation}} \\
    $q$                                   & 91\%          & 4.1          & 0.74          \\
    $q - q^{\mathrm{tgt}}$           & 60\%          & 4.3          & 0.74          \\
    $q - q^{\mathrm{cmd}}$           & \textbf{96\%} & \textbf{3.9} & \textbf{0.77} \\
    \midrule
    \multicolumn{4}{@{}l}{\textit{Controller parameters}} \\
    $k_p \times 2$        & 95\% & 4.0 & 0.72 \\
    $k_p \times 0.5$      & 93\% & 3.8 & 0.74 \\
    $v_{\max} \times 2$   & 95\% & 3.2 & 0.62 \\
    $v_{\max} \times 0.5$ & 76\% & 9.1 & 0.60 \\
    \bottomrule
  \end{tabular*}
\end{table}

We can see that the PCA projection coefficient of $\beta = 0.5$ has the highest success rate, confirming our hypothesis that the PCA can provide a good prior for grasping while allowing the joint space target to refine the PCA pose and compensate for poor re-targeting between the human hand and robot hand. For proprioception, we see that providing the policy with $q - q^{\mathrm{cmd}}$ has the best performance. This is because this number is directly proportional to the torque commanded to the motor, whereas $q - q^{\mathrm{tgt}}$ is computed before it is saturated by the joint velocity limit. We also show that our policy is still able to achieve a high success rate even when varying the stiffness and max velocity on the hand joints post-training. The slower maximum velocity is significantly out of distribution and takes a much longer time to grasp.

\section{Conclusion}
% We present a learned general blind grasping policy that can run on an anthropomorphic robotic hand completely independently of the arm it is attached to.
% We developed an interface that allows the hand to report the grasp confidence to an upstream controller, allowing for recovery from failed grasps and progression in upstream tasks.
% We show that our method generalizes to a large-scale 3D object dataset and also demonstrate 3 real-world examples of arm policies never seen during training communicating with our hand policy to grasp 29 different objects.
% One future work would be integrating a way to allow the hand policy to make small adjustments to the wrist pose through a residual displacement output; this would close the gap to the end-to-end baseline in terms of success rate. 

We present a proprioceptive dexterous grasping policy that acquires
and maintains grasps from hand-local observations and can be paired
with independently designed arm controllers. A bidirectional
arm--hand interface uses the learned grasp score to coordinate
the transition to manipulation and recovery from failed grasps.
We evaluate grasping performance across diverse simulated objects
and show that the policy remains effective under conditions not
encountered during training. Qualitative hardware demonstrations
show successful grasping of 29 real-world objects and reuse of the
same frozen hand policy with three arm behaviors not seen during
training. We also demonstrate reuse across different arm mounting
configurations and robot platforms. Together, these results suggest
that dexterous grasping can be treated as a reusable hand-level skill
rather than being tied to a particular arm controller or perception
pipeline.

\section*{Acknowledgment}{
This work was supported in part by the Technology Innovation Program (RS-2024-00427719) funded by MOTIE, Korea and the KIAT Global Industry Technology Cooperation Center Program (P246800183) and in part by the Advanced Robotics Lab of LG Electronics Co., Ltd.
}

% \addtolength{\textheight}{-12cm}   % This command serves to balance the column lengths
                                  % on the last page of the document manually. It shortens
                                  % the textheight of the last page by a suitable amount.
                                  % This command does not take effect until the next page
                                  % so it should come on the page before the last. Make
                                  % sure that you do not shorten the textheight too much.

%%%%%%%%%%%%%%%%%%%%%%%%%%%%%%%%%%%%%%%%%%%%%%%%%%%%%%%%%%%%%%%%%%%%%%%%%%%%%%%%

%%%%%%%%%%%%%%%%%%%%%%%%%%%%%%%%%%%%%%%%%%%%%%%%%%%%%%%%%%%%%%%%%%%%%%%%%%%%%%%%

%%%%%%%%%%%%%%%%%%%%%%%%%%%%%%%%%%%%%%%%%%%%%%%%%%%%%%%%%%%%%%%%%%%%%%%%%%%%%%%%
% \section*{APPENDIX}

% \section*{ACKNOWLEDGMENT}

%%%%%%%%%%%%%%%%%%%%%%%%%%%%%%%%%%%%%%%%%%%%%%%%%%%%%%%%%%%%%%%%%%%%%%%%%%%%%%%%

% \bibliographystyle{IEEEtran}  % or another style
% \bibliography{tabs/ref}

% \footnotesize{\printbibliography}
\printbibliography

\end{document}